\pdfoutput=1
\documentclass[11pt]{article}

\usepackage[final]{acl}

\usepackage{times}
\usepackage{latexsym}
\usepackage{color}
\usepackage{colortbl}
\usepackage{xcolor}
\usepackage{xspace}
\usepackage{subcaption}
\usepackage{float}
\usepackage{rotating}
\usepackage{hyperref}
\hypersetup{
    colorlinks=true,
    urlcolor=black,
    }
\usepackage[T1]{fontenc}
\usepackage{listings}
\usepackage[utf8]{inputenc}

\usepackage{microtype}
\usepackage{enumitem}
\usepackage{stfloats}
\usepackage{inconsolata}

\usepackage{graphicx}
\usepackage{booktabs}
\newcommand{\gemini}{Gemini\xspace}
\newcommand{\llama}{Llama\xspace}

\title{LLMs for Argument Mining: Detection, Extraction, and Relationship Classification of pre-defined Arguments in Online Comments}
\author{Matteo Guida\textsuperscript{1} \qquad Yulia Otmakhova\textsuperscript{1} \qquad  Eduard Hovy\textsuperscript{2} \qquad Lea Frermann\textsuperscript{1} \\
  \textsuperscript{1}School of Computing and Information Systems, The University of Melbourne \\
  \textsuperscript{2}Language Technologies Institute, Carnegie Mellon University \\
  \texttt{\href{mailto:guida@students.unimelb.edu.au}{guida@student.unimelb.edu.au}}, \\
  \texttt{\{y.otmakhova,eduard.hovy,lea.frermann\}@unimelb.edu.au}
}

\author{Matteo Guida\textsuperscript{1} \qquad Yulia Otmakhova\textsuperscript{1} \qquad  Eduard Hovy\textsuperscript{1} \qquad Lea Frermann\textsuperscript{1} \\
  \textsuperscript{1}School of Computing and Information Systems, \\ The University of Melbourne \\
    \texttt{\href{mailto:guida@students.unimelb.edu.au}{guida@student.unimelb.edu.au}}, \\
    \texttt{\{y.otmakhova,eduard.hovy,lea.frermann\}@unimelb.edu.au}
}

\begin{document}
\maketitle

\begin{abstract}
\textcolor{red}{\it {\bf Content Warning}: This paper discusses examples of harmful language. The authors do not support such content. Reader caution is advised.}\\
Automated large-scale analysis of public discussions around contested issues like abortion requires detecting and understanding the use of arguments. While Large Language Models (LLMs) have shown promise in language processing tasks, their performance in mining topic-specific, pre-defined arguments in online comments remains underexplored. We evaluate four state-of-the-art LLMs on three argument mining tasks using datasets comprising over 2,000 opinion comments across six polarizing topics. Quantitative evaluation suggests an overall strong performance across the three tasks, especially for large and fine-tuned LLMs, albeit at a significant environmental cost. However, a detailed error analysis revealed systematic shortcomings on long and nuanced comments and  emotionally charged language, raising concerns for downstream applications like content moderation or opinion analysis. Our results highlight both the promise and current limitations of LLMs for automated argument analysis in online comments.\footnote{Our code, data and prompts can be found at: \url{https://anonymous.4open.science/r/llm-for-arg-min-35DA}}
\end{abstract}

\section{Introduction}
Online discourse on social media or in discussion fora on complex controversial topics brings both challenges and opportunities for understanding the formation and spread of opinions, and their expression through arguments, at scale. Automatic analysis of public debate is crucial for tracking how opinions form and spread, identifying the evidence supporting different viewpoints, and evaluating the quality of public discourse~\citep{stede2019argumentation}.

Accordingly, a rich body of work on computational argument mining and understanding has emerged which includes the detection of argumentative discourse units in texts~\citep{habernal_argumentation_2017, hidey_analyzing_2017}, the relationships of these units (in terms of attack and support, \citep{carstens_towards_2015, Ruiz_Dolz_2021} and the identification of use cases of pre-defined arguments in heterogeneous texts~\citep{boltuzic_back_2014, hasan_why_2014, levy2014context}.
This latter approach enables researchers to abstract away from individual expressions by aggregating them into pre-defined argument types, thereby facilitating the analysis of broader and recurring argumentation patterns that would be difficult to capture through `bottom-up' argument mining.

\begin{figure}
\includegraphics[width=\columnwidth]{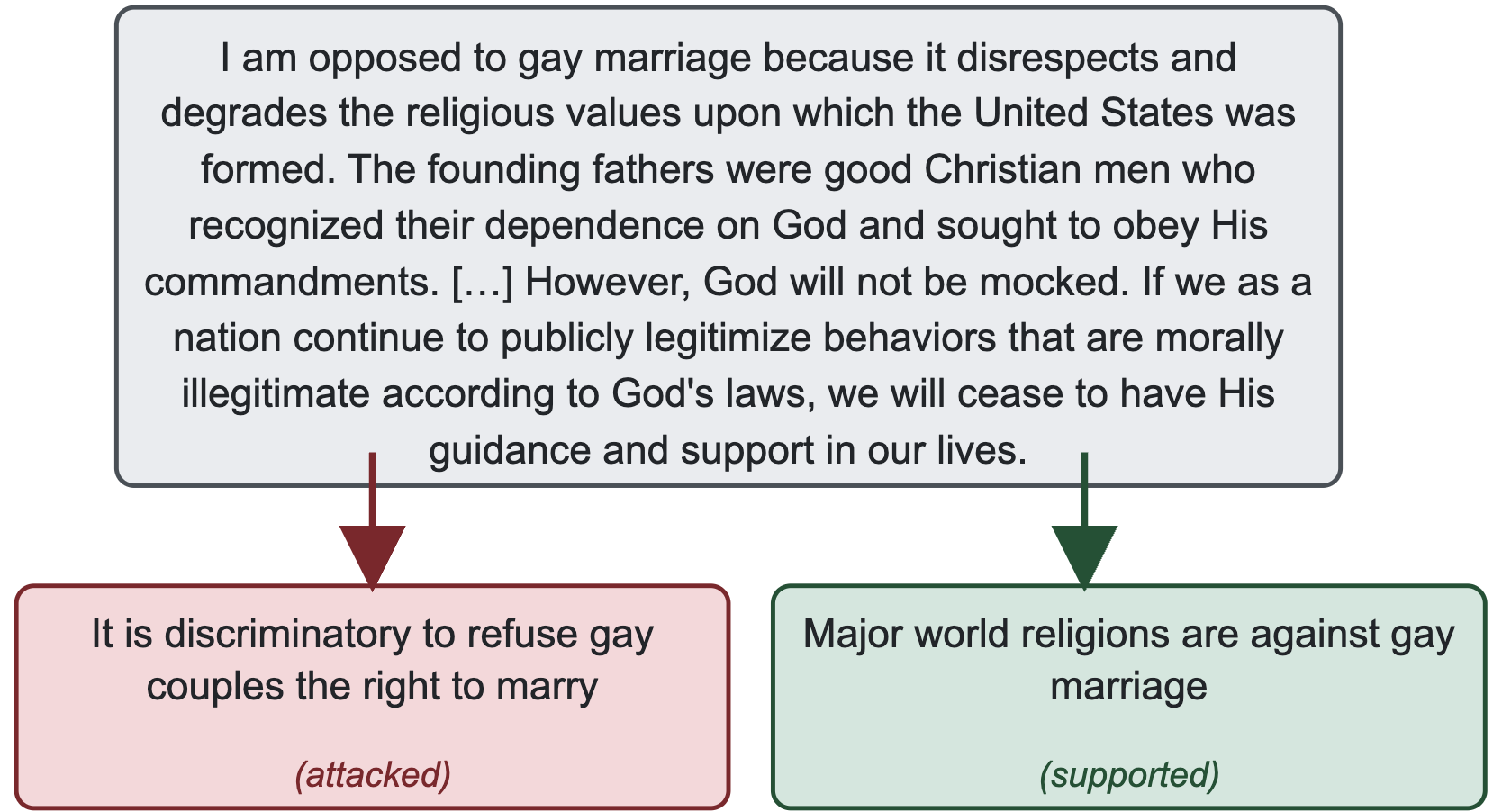}
\caption{An online comment (top) which makes use of two pre-defined arguments (red and green boxes). The comment attacks A1 (left) and supports A2 (right).}
\label{fig:example_comment}
\end{figure}

Here, we build on this latter approach. Specifically, we start out with a controversial \textit{topic} ("Legalisation of Abortion"), paired with a set of pre-defined \textit{arguments}\footnote{In this paper, we use the term \textit{argument} to refer to "a general, concise statement that directly supports or contests the given topic", following ~\citealp[pg.1489]{levy2014context}.} which can be either in favour of ("Abortion is a woman's right"), or against the topic ("Abortion kills a life"). Our goal is to identify usages of these arguments in online comments. A comment can make use of an argument by either \textit{supporting} or \textit{attacking} it (Figure~\ref{fig:example_comment}). 

\begin{figure}
\includegraphics[width=\columnwidth]{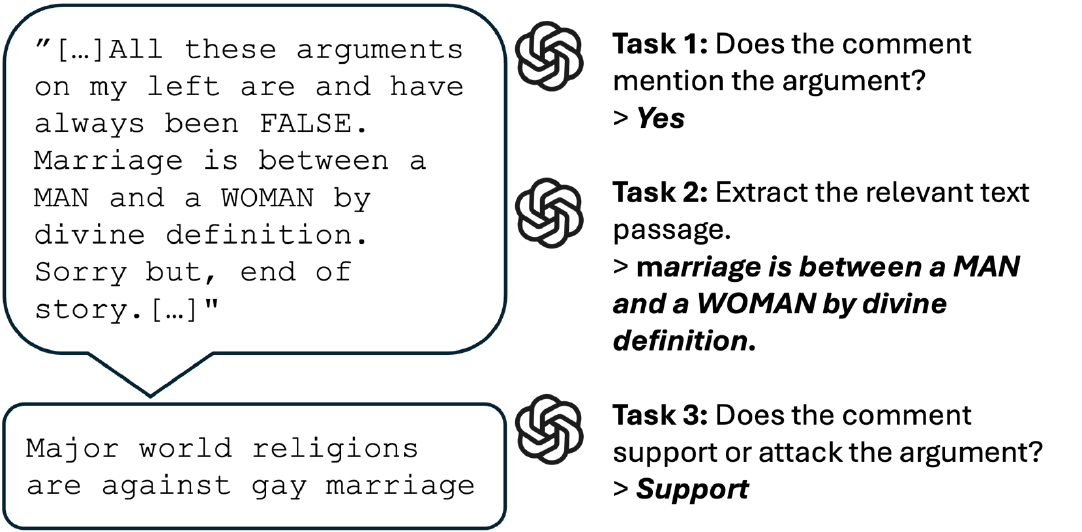}
\caption{A comment (top, left) and pre-defined argument (bottom, left). We predict whether the comment makes use of the argument (Task 1), where it mentions the argument (Task 2) and whether it supports or attacks the argument (Task 3).}
\label{fig:page1}
\end{figure}

Correspondingly, we formulate three tasks to disentangle models' performance: identify whether an argument is used in a comment (Task 1); extract the span of usage (Task 2); and assess whether the argument is supported or attacked in the comment (Task 3). This is illustrated in Figure~\ref{fig:page1}. While these tasks are not new (cf.,~Section 2), taken together they provide a comprehensive picture of model performance. 

On this basis, we make two empirical contributions. First, we inspect the ability of large-language models (LLMs) to identify pre-defined arguments in noisy online comments. With the increased performance and adoption of LLMs, and large-scale opinion analysis in social media being a conceivable use-case, to the best of our knowledge LLMs have not yet been systematically tested on these tasks using a set of topic-specific, pre-defined arguments. Second, given the sensitive nature of the task and consequential importance to avoid systematic bias in model performance, we conduct a detailed qualitative and quantitative error analysis of model outputs.

To assess LLMs on the proposed tasks, we utilize datasets of over 2,000 opinion comments spanning six polarizing topics~\citep{boltuzic_back_2014, hasan_why_2014}. For each topic, a set of pre-defined arguments has been identified, and comments were annotated for the presence and usage (support vs attack) of arguments. We experiment with four state-of-the-art LLMs comprising open and closed-source models of varying sizes. Our findings are two-fold: first, fine-tuned LLMs outperform both prompted LLMs and traditional fine-tuned models (RoBERTa) on argument detection and extraction tasks, although at a significant environmental cost. Second, our error analysis exposes systematic weaknesses: LLMs frequently over-predict arguments in comments using strong and emotional language, struggle to distinguish the implicit and explicit use of arguments, and perform worse on longer, more nuanced comments. These patterns suggest that while LLMs show promise for argument mining using pre-defined arguments, their current limitations could lead to biased analyses in applications like public opinion analysis or content moderation.
% In sum, the contributions of this paper are:
% \begin{itemize}
% \item We conduct the first comprehensive evaluation of LLMs for argument mining tasks with pre-defined arguments in opinion comments
% \item We present empirical evidence that LLMs do not uniformly outperform smaller fine-tuned models, especially for span extraction
% \item We discuss the ethical implications for deploying LLMs in argument mining systems
% \end{itemize}

\section{Related Work}

\paragraph{Argument Mining} A vast body of work has studied argumentation from theoretical and empirical perspectives, adopting  an open-domain, bottom-up approach to identify argumentative units directly from unstructured text. Early research focused on automatically identifying arguments (or premises) and conclusions (or claims) in opinionated texts such as essays or online discussions~\citep{habernal_argumentation_2017, hidey_analyzing_2017, feng_classifying_2011, stab-gurevych-2017-parsing}. Other work examined the interaction of these components, as premises supporting or attacking a claim~\citep{cocarascu_identifying_2017, carstens_towards_2015, Ruiz_Dolz_2021}. These tasks are often addressed jointly through structured prediction models~\citep{egawa_corpus_2020, stab-gurevych-2017-parsing}.

% Argument structure analysis faces significant challenges, as the identification of argument components (e.g., claims, premises) is subjective, with no clear linguistic consensus on their precise definition or characteristics~\citep{daxenberger-etal-2017-essence}, making evaluation difficult~\citep{mestre2022benchmark}. 

%Furthermore, most work identifies arguments on an ad-hoc, document-level basis without mapping them back to recurring argument patterns—broader, topic-specific arguments that recur across discussions. This limits their utility for analyzing discourse at scale. To address this, we test LLMs on a structured task: identifying pre-defined argument templates (e.g., "\textit{A fetus is not a human yet, so it’s okay to abort}") in online comments and classifying their usage (support/attack). 

\paragraph{Argument Mining with pre-defined Arguments} Another line of research focuses on identifying pre-defined arguments -- typically sourced from debate platforms -- in unstructured text. In works on argument search, for instance, such arguments are retrieved and ranked from such platforms in response to user queries \citep{al2017building, stab2018argumentext}. Beyond argument search and ranking, \citet{levy2014context} automatically detected claims from Wikipedia articles that were relevant to a set of pre-defined arguments. 

Similarly, key point analysis (KPA) identifies lists of "key points" that summarize arguments about a variety of topics \citep{bar-haim-etal-2020-quantitative, barhaim2020argumentskeypointsautomatic} and is thus similar in flavour to our Task 1. These KPA datasets are based on crowd-sourced arguments with a strict length limitation (210 characters max as opposed to a {\it median} 480 characters in the data we use -- see Table~\ref{tab:text-length-stats} and Table~\ref{tab:comments-examples} for complete statistics and examples), and with crowd-sourced associated key points. While evaluation on KPA data sets is a worthwhile avenue for future work, in this paper we focus (a) on datasets that support all three evaluation tasks, which allows for a comprehensive evaluation of LLMs, and (b) real-world online commentary, which is more representative of natural, varied, and "heated" discussions, thus potentially harder for the model to understand.

To do so, we build on~\citet{boltuzic_back_2014} and \citet{hasan_why_2014}, who developed datasets which labelled opinion comments on divisive issues (like abortion) with the presence and usage of carefully crafted pre-defined issue-related arguments from online debate platforms (details in Section~\ref{data}). The original works trained SVMs and Maximum Entropy models, respectively, on selected subsets of our proposed tasks.

\paragraph{Argument Mining with Large Language Models} 
LLMs have caused substantial performance gains across argument mining tasks ranging from argument extraction~\citep{de_wynter_id_2024}, understanding~\citep{gorur_can_2024,otiefy_exploring_2024}, and quality assessment~\citep{van_der_meer_hyena_2022}. However, for tasks like argument generation and persuasiveness \citep{hinton2023persuasive} and argumentative fallacy identification~\citep{ruiz2023detecting} results were mixed. Similarly, cross-task review papers on argument mining have reported mixed results~\citep{chen_exploring_2024,alsubhi_arabig5_2023,ruiz-dolz_overview_2024}.
% Other work has analyzed the ability of LLMs to detect persuasive arguments~\cite{Rescala2024CanLM} and evaluate argument quality~ \cite{Mirzakhmedova2024AreLL}.
 
% Existing reviews of LLM performance on argument mining tasks drew inconsistent conclusions. The most comprehensive systematic review of LLMs performance in argument mining and argument generation tasks to date is~\citet{chen_exploring_2024}. The authors performed zero-shot and k-shot experiments using GPT-3.5, Flan-T5 and Llama2 models on a variety of argument mining tasks (claim detection, evidence detection, stance detection, evidence classification), as well as argument generation and summarization. Their results highlight decent performance on binary classification tasks, but worse with more complex, multi-label classification tasks. Other reviews, however, showed that competitive LLMs (including GPT-4) did not suprise domain-specific fine-tuned BERT-family models~\citep{alsubhi_arabig5_2023,ruiz-dolz_overview_2024}.

We complement this line of evaluation with the first comprehensive assessment of LLMs to detect and understand pre-defined arguments in opinion comments (but see~\citet{gorur_can_2024} for a study specific to relation classification). We systematically assess fine-tuned, and few-shot LLMs on all three defined tasks and conduct detailed qualitative and quantitative error analyses.
% Given the rapid adoption and documented biases of LLMs~\citep{tlaie2024exploring, Pit2024WhoseSA}, a systematic assessment is urgent: can LLMs, via few-shot learning or fine-tuning, robustly automate these tasks? Our work fills this gap, providing insights to foster discussion about deploying LLMs for scalable argument retrieval.

\section{Methodology}

\subsection{Data}
\label{data}
Our study builds on prior research in natural language processing, particularly works that intersected curated arguments from online debate platforms with large-scale online discussions. 

The \textbf{COMARG dataset}:~\citet{boltuzic_back_2014}  manually annotated  373 comments from the discussion platform \textit{Procon.org} with a pre-defined list of arguments retrieved from \textit{Idebate.org}. It encompasses two topics:  the legalisation of gay marriage (GM) and the inclusion of the phrase "Under God" in the U.S. Pledge of Allegiance (UGIP). GM comments were labeled for the presence of three arguments in favor (Pro) and four arguments against the topic (Con), while the UGIP topic featured three Pro and three Con arguments. Each attested comment-argument pair was further classified based on whether the comment explicitly supported, implicitly supported, explicitly attacked or implicitly attacked the argument. Inter-annotator agreement was moderate, and the final labels were decided by majority vote, excluding all cases where no majority was reached.
% Although the authors reported moderate inter-annotator agreement, the task proved highly subjective and challenging. Furthermore, inconsistencies were found in the original dataset, where certain comments lacked annotations for some arguments. To address this, we included all arguments in our dataset and automatically labeled any unannotated relationships as "makes no use" to maintain consistency for evaluation.

The \textbf{YRU dataset}:~\citet{hasan_why_2014} sourced 1900 comments from \textit{createdebate.com}, covering four topics: abortion (AB), gay rights (GR), legalization of marijuana (MA), and the Obama presidency (OB). For each topic, annotators identified a set of 6-9 arguments each supporting and opposing the topic. The data set was originally developed for the task of argument extraction, i.e., manually labeled with spans of text that employed a specific argument. Annotator agreement on this labelling task was reported as moderate to high, and disagreements were resolved through discussion.
Table~\ref{proconargs} in the Appendix lists all arguments for the six topics across both datasets. 

% \begin{table}[ht]
% \centering
% \begin{tabular}{l|p{2.5cm}|p{2.5cm}}
% \hline
% \textbf{Dataset} & \textbf{Size} & \textbf{Topics} \\ \hline
% COMARG & 373 texts & Gay Marriage, Under God in Pledge \\ \hline
% YRU & 1901 texts & Abortion, Gay Rights, Marijuana, Obama \\ \hline
% \end{tabular}
% \caption{Description of the COMARG and YRU datasets.}
% \label{tab:dataset_description}
% \end{table}

\subsection{Task Definitions}
We assess our models on three argument mining tasks designed to test their abilities to {\it detect}, {\it extract}, and {\it understand the use of} arguments in online comments.

\paragraph{Task 1: Binary Argument Detection}     Given an argument \( A \) and a comment \( C \), the task is to classify, in binary fashion, whether \( C \) makes use of \( A \). We run this task on both YRU and COMARG, across all six topics. 

\paragraph{Task 2: Argument Span Extraction} 
    Given an argument \( A \) and a comment \( C \), the goal is to extract the span within \( C \) that expresses \( A \). Only the COMARG dataset comes with manually annotated argument spans, so we evaluate this task over the four COMARG topics.

\paragraph{Task 3: Argument Relationship Classification}
    Given an argument \( A \) and a comment \( C \), we determine the relationship between \( A \) and \( C \) as $C$ either attacking or supporting $A$. We consider two formulations of this task: either a {binary} classification as support or attack; or a 4-way classification distinguishing between explicit/implicit support for or an explicit/implicit attack of an argument. Only the YRU dataset labels the type of usage of an argument, so we evaluate relation classification over the two topics in this dataset.

\subsection{Data Pre-Processing}
For binary argument detection (Task 1) we pre-processed the original datasets to conform to support a binary classification task. For the COMARG dataset we consider all comment-argument pairs labeled as exhibiting any form of argumentative relationship as present (1). The data contained an explicit label of `makes no use of an argument', which we reuse as our negative (not present) label (0). The YRU dataset is annotated for arguments on the sentence level. We project these labels to the comment-level, and consider them as present (1). All arguments not identified in any sentence were labeled as not present (0). 

For the span extraction (Task 2), we only considered the labels present in the original YRU dataset and the manually annotated spans in the comment. Finally,
for the argument relationship classification (Task 3), we treated the data in the COMARG dataset differently for the two subtasks. In subtask 3a we conflated the original labels in a binary fashion, only aiming at identifying whether the comment supports or attacks the argument. For subtask 3b we considered the original scale of implicit/explicit support and attack, we thus left the original 4-way labeling unaltered.

\subsection{Models}
We selected four Large Language Models (LLMs) from different model families, spanning one open-source -- Llama3-8b-Instruct \citep{dubey2024llama} -- and three proprietary models: GPT4o-mini and GPT-4o \citep{achiam2023gpt}, and Gemini1.5-Flash \citep{reid2024gemini}.
% We tested these models' performance using three distinct prompting strategies for each task: zero-shot, one-shot and five-shot prompting. 
We followed established practices to minimize non-deterministic behavior and output variability \citep{zhang2023investigating, meng2023enhancing}, i.e. setting the temperature to 0 and the top\_p parameter to 1 \citep{liu2023empirical, brown2023understanding}.
%To minimize non-deterministic behavior in the models' outputs, we followed established practices for controlling generation parameters \citep{zhang2023investigating, meng2023enhancing}. Specifically, we set the temperature parameter to 0 and the top\_p parameter to 1, as these settings have been shown to reduce output variability and increase reproducibility \citep{liu2023empirical, brown2023understanding}. 
% This approach aligns with previous studies demonstrating that lower temperature values lead to more consistent and deterministic outputs \citep{kumar2024systematic}.
\footnote{For Llama3-8b-Instruct, we also set the top\_k parameter to 1. GPT4o-mini and Gemini1.5Flash do not feature manual configuration of this parameter.} 

\paragraph{Prompts} In preliminary experiments, we varied our prompts along three key dimensions: structure (unstructured vs. structured step-by-step instructions), specificity (varying level of detail on task requirements and constraints), and role assignment (including/excluding the specific assignment of a role such as ``you are an expert in argument analysis''). For argument detection (Task 1), a structured prompt with detailed instructions but without role assignment performed best. For both span extraction (Task 2) and argument relationship classification (Task 3), prompts that combined structured step-by-step instructions with explicit role assignment achieved superior performance. These optimized prompts were used for all subsequent experiments.\footnote{The full prompts are released in our repository.}

Each task was attempted as zero-shot, 1-shot and 5-shot. To assess the impact of chosen examples, each few-shot experiment was run five times with randomly sampled, non-overlapping instruction examples. We manually verified that examples were instructive, and that the five-shot example set covered all classes.

\paragraph{RoBERTa Baselines} 
We fine-tuned one RoBERTa model \citep{liu2019roberta} for each task, by combining all the available data across topics. The relatively small number of samples for individual topics renders topic-wise fine-tuning infeasible. 
% For Task 1, we merged the COMARG and YRU datasets, utilizing the processed binary labels. Task 2 involved the convergence of GM and UGIP data from the COMARG dataset, necessitating two distinct RoBERTa models for its respective subtasks. In subtask 2a, we consolidated the labels representing varying degrees of support/attack relationships, while subtask 2b preserved the polarity distinctions (implicit/explicit) within the support/attack labels.

For the classification tasks, we concatenated each comment-argument pair using the [SEP] token as a delimiter. We randomly split the data into five stratified folds for cross-validation, ensuring a balanced label distribution in each split. Each model was trained for 3 epochs with a batch size of 16. 
For the span extraction task, we formatted the data equivalent to extractive question-answer tasks, where arguments serve as ``question'', and relevant spans as the ``answer'' to be extracted. We fine-tuned a RoBERTa model on this data using the QuestionAnsweringModel from SimpleTransformers\footnote{\url{https://simpletransformers.ai/docs/qa-model/}} again with five fold stratified cross validation, training for a total of 10 epochs and with a batch size of 16.\footnote{Information about the parameters are reported inc  Appendix~\ref{robfinetun}.}

\paragraph{LLM Fine-tuning} To disentangle the effect of fine-tuning from model size, we also fine-tune one of our LLMs. For Llama3-8b-Instruct we performed parameter-efficient fine-tuning using low-rank adaptation (LoRA) \cite{hu2021lora}, with cross-validation on five stratified folds. The details of hyperparameters and training protocol are provided in Appendix~\ref{app:lora}. We include fine-tuned Llama only for the argument detection task and the argument extraction task, because the fine-tuned RoBERTa for the relationship classification task was widely outperformed by all LLMs in the prompting setup.

\begin{table}[t]
  \begin{small}
  \setlength{\tabcolsep}{0.39em}
    \begin{tabular}{l|ccccccc}
    \toprule
    Model & \multicolumn{1}{c}{GM} & \multicolumn{1}{c}{UG} & \multicolumn{1}{c}{AB} & \multicolumn{1}{c}{GR} & \multicolumn{1}{c}{MA} & \multicolumn{1}{c}{OB} & \multicolumn{1}{c}{Comb} \\\midrule
    {\bf Majority} & 0.40 & 0.41 & 0.47 & 0.47 & 0.46 & 0.48 & 0.44 \\
    {\bf RoBERTa} &  &  &  &  &  &  & 0.61 \\\midrule
    \multicolumn{8}{c}{\bf Zero shot} \\\midrule
    {\bf Gemini1.5-f} & 0.79 & 0.73 & 0.73 & 0.67 & 0.66 & 0.67 & 0.72 \\
    {\bf GPT4o} & 0.76 & \textbf{0.75} & \textbf{0.81} & 0.72 & \textbf{0.68} & 0.66 & 0.68 \\
    {\bf GPT4o-m} & 0.75 & 0.74 & 0.76 & 0.67 & 0.66 & 0.67 & 0.69 \\
    {\bf Llama3} & 0.69 & 0.65 & 0.65 & 0.65 & 0.63 & 0.63 & 0.65 \\\midrule
    \multicolumn{8}{c}{\bf One shot} \\\midrule
    {\bf Gemini1.5-f} & \textbf{0.80} & \textbf{0.75} & 0.74 & 0.68 & 0.67 & 0.67 & 0.72 \\
    {\bf GPT4o} & 0.75 & 0.73 & 0.79 & \textbf{0.73} & 0.65 & \textbf{0.68} & 0.73 \\
    {\bf GPT4o-m} & 0.78 & 0.63 & 0.75 & 0.67 & 0.67 & 0.67 & 0.70 \\
    {\bf Llama3} & 0.63 & 0.63 & 0.62 & 0.63 & 0.59 & 0.60 & 0.61 \\\midrule
    \multicolumn{8}{c}{\bf Five shot} \\\midrule
    {\bf Gemini1.5-f} & \textbf{0.80} & 0.74 & 0.73 & 0.67 & 0.67 & 0.67 & 0.73 \\
    {\bf GPT4o} & 0.76 & 0.72 & 0.76 & 0.71 & 0.66 & \textbf{0.68} & 0.71 \\
    {\bf GPT4o-m} & 0.75 & 0.63 & 0.75 & 0.68 & \textbf{0.68} & 0.67 & 0.70 \\
    {\bf Llama3} & 0.60 & 0.62 & 0.61 & 0.63 & 0.59 & 0.59 & 0.60 \\\midrule
    \textbf{Llama3 FT} &  &  &  &  &  &  & \textbf{0.76} \\
    \bottomrule
    \end{tabular}
  \end{small}
    \caption{Results for binary argument detection (Task 1) for six topics and the combined data set (final column) as macro-averaged F1. We report a majority baseline, and fine-tuned RoBERTa and fine-tuned Llama3 (Llama3 FT) on the combined data only. The best F1 scores per data set are bolded. 1-shot and 5-shot results are averaged over five runs. The majority baseline is defined as predicting the most frequent class in the training data.}
  \label{task1results}
\end{table}

\begin{table}
  \begin{small}
    \begin{tabular}{l|ccccc}
    \toprule
    Model & AB & GR & MA & OB & Comb\\ \midrule
    {\bf RoBERTa} & & & & & 0.44 \\\midrule
    & \multicolumn{5}{c}{\bf Zero shot} \\ \midrule
    {\bf Gemini1.5-flash} & 0.42 & 0.41 & 0.37 & 0.38 & 0.40 \\
    {\bf GPT4o} & 0.31 & 0.32 & 0.30 & 0.32 & 0.31 \\
    {\bf GPT4o-m} & 0.28 & 0.29 & 0.27 & 0.25 & 0.27 \\
    {\bf Llama3} & 0.29 & 0.33 & 0.27 & 0.28 & 0.29 \\\midrule
    & \multicolumn{5}{c}{\bf One shot} \\ \midrule
    {\bf Gemini1.5-flash} & 0.46 & 0.46 & 0.43 & 0.47 & 0.46 \\
    {\bf GPT4o} & 0.36 & 0.41 & 0.37 & 0.41 & 0.39 \\
    {\bf GPT4o-m} & 0.35 & 0.38 & 0.37 & 0.36 & 0.37 \\
    {\bf Llama3} & 0.36 & 0.42 & 0.37 & 0.41 & 0.39 \\\midrule
    & \multicolumn{5}{c}{\bf Five shot} \\ \midrule
    {\bf Gemini1.5-flash} & \textbf{0.50} & \textbf{0.51} & \textbf{0.48} & \textbf{0.55} & 0.51 \\
    {\bf GPT4o} & 0.44 & 0.48 & 0.42 & 0.47 & 0.45 \\
    {\bf GPT4o-m} & 0.43 & 0.46 & 0.42 & 0.43 & 0.44 \\
    {\bf Llama3} & 0.48 & 0.50 & 0.43 & 0.50 & 0.48 \\\midrule
    {\bf Llama3 FT} & & & & & \textbf{0.54} \\
    \bottomrule
    \end{tabular}
  \end{small}
  \centering\caption{Results for Argument Extraction (Task 2) for the four topics in the YRU data set and the combined data set (final column) as Rouge-L. Models as in Table~\ref{task1results}. The best Rouge-L scores per data set are bolded. 1-shot and 5-shot results are averaged over five runs with different examples.}
  \label{task3results}
\end{table}

\section{Results}
We first present the quantitative results of our four LLMs and baselines across tasks, followed by a detailed error analysis. Overall, we find that (1)~fine-tuned Llama outperformed all other models in detecting and extracting arguments; (2)~larger LLMs generally outperformed smaller models and are more robust to different few-shot examples (exhibiting smaller variance); (3)~that instruction examples (one- or five-shot) do not necessarily lead to enhanced performance; and (4)~that the {\it detection} of arguments in comments (Task 1) is more challenging for LLMs than binary relationship classification (Task 3), which calls for caution with and future research on automated argument extraction in online discussion.

\subsection{Task 1: Binary Argument Detection}
We test four models (\llama, GPT4o, GPT4o-mini, \gemini) in 0-, 1-, and 5-shot settings across six different topics on predicting whether a given argument is stated in a comment or not. Results in Table~\ref{task1results} show that all LLMs outperform the baselines, and that the fine-tuned Llama3 performs best overall.\footnote{For task 1, the F1 SDs of the fine-tuned LLM range from ±0 to ±0.01, indicating robustness.} Among the prompt-based models, the largest variants (GPT4o and Gemini) outperform their smaller counterparts. We observe a strong variance across topics, with abortion (AB) and gay marriage (GM) performing best. Finally, and perhaps counterintuitively, we do not observe consistent improvement with more examples. The standard deviation (std) across five model runs for few-shot experiments was ±0.01 to ±0.02 for larger models, indicating high robustness to varying inputs, while smaller models showed slightly higher std, ±0.02 to ±0.03, especially in 1-shot settings.

\subsection{Task 2: Argument Extraction}
Here, we tasked models with identifying the exact span of text in which an argument is being used. We report the ROUGE-L scores~\cite{lin2004rouge} between predicted and golden spans.

Results in Table~\ref{task3results} reveal that, similar as in Task~1, the fine-tuned Llama3 outperformed all other models.\footnote{With F1 standard deviations ranging from 0.01 to 0.015 across the folds, indicating stability} In prompting experiments, 5-shot \gemini consistently performs best. We observe a consistent improvement with exposure to more examples in the task instruction. We posit that this is due to the extractive nature of the task, which is more challenging for LLMs out-of-the-box compared to classification (Task 1). Most interestingly, we observe that most LLMs outperform the RoBERTa baseline only in the 5-shot setting on the combined data set, and the gap between non-fine tuned LLMs and RoBERTa is small (with the exception of 5-shot \gemini).
Larger models (Gemini, GPT4o) show low std (±0.01 to ±0.03), while smaller models (GPT4o-mini, Llama) exhibit slightly higher std (±0.02 to ±0.05), especially in 5-shot settings.

While ROUGE-L evaluates strict lexical overlap, it disproportionately penalizes extracted spans that use different wordings to express the same point as in the golden spans. For example, for the argument "\textit{Gay people should have the same rights as straight people}", a gold span "\textit{Its not our job to tell people what they should do}" and a predicted span "\textit{Personally, I think love is equal, whether is in the form of a man and a woman, a man with a man, or a woman with a woman}" are both expressions of the given argument, but achieve a ROUGE-L score of only 0.08 due to low lexical overlap, ignoring their semantic affinity. To assess this, we additionally computed BERTScore~\citep{zhang2020bertscoreevaluatingtextgeneration}, which computes token-level semantic similarity using BERT contextual embeddings, for the best-performing model (Gemini) averaged over all data sets. Across splits, BERTScores are consistently high (mean F1=0.87–0.91). While BERTScore is known to over-estimate extractive performance of models, and should not be used as the sole metric in a task like argument understanding where subtle differences in wording have large effects, a comparison of both metrics and manual inspection suggests that the ROUGE-L scores are a lower-bound of true model performance.
% strict ROUGE-L and lenient BERTScores provides a more complete picture of performance. , confirming consistent high semantic similarity between gold and extracted spans and indicating that LLMs often extract spans that are semantically aligned with, yet lexically distinct from, the gold span. 

\begin{table}
\centering
  \begin{small}
  \setlength{\tabcolsep}{0.5em}
    \begin{tabular}{l|ccc|ccc}
    \toprule
     & \multicolumn{3}{c}{Binary} & \multicolumn{3}{c}{Scale} \\
     Model & GM  & UG  & Comb  & GM  & UG  & Comb  \\
    \midrule
    {\bf Majority} & 0.39 & 0.37 & 0.38 & 0.14 & 0.37 & 0.25 \\
    {\bf RoBERTa} &  &  & 0.39 &  &  & 0.15 \\
    \midrule
    \multicolumn{7}{c}{\bf Zero shot} \\
    \midrule
    {\bf Gemini1.5-f} & 0.92 & \textbf{0.96} & 0.94 & 0.55 & 0.59 & 0.57 \\
    {\bf GPT4o} & 0.94 & \textbf{0.96} & \textbf{0.95} & 0.56 & \textbf{0.61} & 0.58 \\
    {\bf GPT4o-m} & 0.77 & 0.91 & 0.84 & 0.40 & 0.40 & 0.40 \\
    {\bf Llama3} & 0.83 & 0.78 & 0.80 & 0.34 & 0.45 & 0.39 \\
    \midrule
    \multicolumn{7}{c}{\bf One shot} \\
    \midrule
    {\bf Gemini1.5-f} & \textbf{0.93} & 0.90 & 0.91 & \textbf{0.57} & \textbf{0.61} & \textbf{0.59} \\
    {\bf GPT4o} & 0.71 & 0.86 & 0.78 & 0.40 & 0.40 & 0.40 \\
    {\bf GPT4o-m} & 0.65 & 0.81 & 0.73 & 0.37 & 0.38 & 0.37 \\
    {\bf Llama3} & 0.55 & 0.73 & 0.64 & 0.30 & 0.30 & 0.30 \\
    \midrule
    \multicolumn{7}{c}{\bf Five shot} \\
    \midrule
    {\bf Gemini1.5-f} & \textbf{0.93} & \textbf{0.96} & 0.94 & \textbf{0.57} & \textbf{0.61} & \textbf{0.59} \\
    {\bf GPT4o} & 0.68 & 0.92 & 0.80 & 0.40 & 0.40 & 0.40 \\
    {\bf GPT4o-m} & 0.64 & 0.86 & 0.75 & 0.37 & 0.37 & 0.37 \\
    {\bf Llama3} & 0.54 & 0.74 & 0.64 & 0.29 & 0.29 & 0.29 \\
    \bottomrule
    \end{tabular}
  \end{small}
  \caption{Results for Argument Relationship Classification (Task 3) showing F1 scores. Left: binary classification (support vs attack); Right: 4-way classification (explicit/implicit support/attack). The best F1 scores per data set are bolded. 1-shot and 5-shot results are averaged over five runs. The majority baseline is defined as predicting the most frequent class in the training data.}
  \label{task2results}
\end{table}

\subsection{Task 3: Argument Relationship Classification}
Given a comment and an argument featured in the comment, we ask models whether the argument is \textit{supported} or \textit{attacked} in the comment, either in a \textbf{binary} fashion, or on a 4-way \textbf{scale} (explicitly/implicitly supports; explicitly/implicitly attacks). 
Focusing on the binary task (Table~\ref{task2results}, left) we observe that the two largest models (Gemini and GPT4o) consistently perform best, achieving almost perfect results. Exposure to examples does not improve performance and, in fact, substantially decreases results for GPT4-mini and Llama3. 
We observe a substantial performance decrease when moving to the 4-way classification task (Table~\ref{task2results}, right), with the larger LLMs again performing best. 
The F1 std for the models show that Gemini1.5-f indicates low variability (std ±0.02), while GPT-4o-m and GPT-4o have substancial variability (std ±0.03 to ±0.16), and Llama3 shows even higher variability (std ±0.07 to ±0.10).

{RoBERTa fails on this task, barely outperforming the Majority baseline, due to the small number of instance per label. This is supported by the fact that RoBERTa achieves better results on the binary classification than on the 4-way classification task, where class merging increases the number of examples per category.}

Interestingly, performance across models was higher in the binary version of Task 3 than Task~1. In other words, models do better at identifying whether a comment \textit{supports or attacks} a given argument than at detecting whether a comment \textit{uses} the argument. The models benefited from examples uniformly only for argument extraction (Task 2), but not in the classification tasks. Consistently, a fine-tuned RoBERTa model performed competitively with the LLMs on Task 2. 

%Overall, we conclude that there is substantial room for improvement in LLM argument detection and interpretation for all presented task with the exception of binary argument relation classification. 

% \begin{table*}
%   \begin{small}
%   \setlength{\tabcolsep}{0.4em}
%     \begin{tabular}{l|ccc|ccc}
%     \toprule
%     Model & \multicolumn{3}{c|}{GM} & \multicolumn{3}{c}{UG}  \\
%     & P & R & F1 & P & R & F1  \\\midrule
%     {\bf Majority (scale)}  & 0.10 & 0.25 & 0.14  & 0.29 & 0.50 & 0.37 \\\midrule
%     {\bf RoBERTa (scale)}  & 0.16 & 0.40 & 0.23  & & &  \\\midrule
%     & \multicolumn{6}{c}{\bf Zero shot - scale}\\\midrule
%     {\bf Llama3-8b} & 0.24 & 0.25 & 0.24 & 0.39 & 0.42 & 0.40 \\
%     {\bf GPT4o-m} & 0.24 & 0.23 & 0.24 & 0.23 & 0.42 & 0.29 \\
%     {\bf Gemini1.5-flash} & 0.46 & 0.43 & 0.44 & 0.36 & 0.42 & 0.39 \\\midrule
%     & \multicolumn{6}{c}{\bf One shot - scale}\\\midrule
%     {\bf Llama3-8b} & 0.25 & 0.27 & 0.26 & 0.38 & 0.37 & 0.37 \\
%     {\bf GPT4o-m} & 0.22 & 0.22 & 0.22 & 0.31 & 0.39 & 0.34 \\
%     {\bf Gemini1.5-flash} & 0.43 & 0.42 & 0.42 & 0.54 & 0.45 & 0.49 \\\midrule
%     & \multicolumn{6}{c}{\bf Five shot - scale}\\\midrule
%     {\bf Llama3-8b} & 0.33 & 0.30 & 0.32 & 0.37 & 0.31 & 0.33 \\
%     {\bf GPT4o-m}& 0.24 & 0.25 & 0.25 & 0.28 & 0.35 & 0.31 \\
%     {\bf Gemini1.5-flash} & 0.39 & 0.43 & 0.41 & 0.41 & 0.36 & 0.38 \\
%     \bottomrule
%     \end{tabular}
%   \end{small}
%   \centering\caption{Task 2 -- scale}
%   \label{task2results}
% \end{table*}

\begin{table*}
\centering
\begin{small}
\setlength{\tabcolsep}{0.4em}
\begin{tabular}{p{1.6\columnwidth}|p{0.3\columnwidth}|p{0.09\columnwidth}}
\toprule
\textbf{Comment} & \textbf{Argument} & \textbf{Topic} \\%& \textbf{Language} \\
\midrule
I think every woman and anyone that's for abortions, that has a voluntary abortion should have every reproduction organ removed from their body [...] & Unwanted babies are ill-treated by parents and/or not always adopted  & AB \\%& Strong\\
\midrule
Obama is another Hitler. There is not an ounce of capitalism or freedom in him.  Why won't anybody in the media talk against him?  Its because of the fairness doctrine. You're not allowed to speak against him.  Stop listening to the liberal media. & Not eligible as a leader & OB \\%& Strong
\bottomrule
\end{tabular}
\end{small}
\caption{Representative examples of false positive (FPs) predictions in Task 1, where the model falsely detected an argument in a comment. FPs often occur for comments that use strong/emotional language.}
\label{tab:error_analysis_task1}
\end{table*}

\begin{figure}[t]
    \centering
    \includegraphics[width=\linewidth]{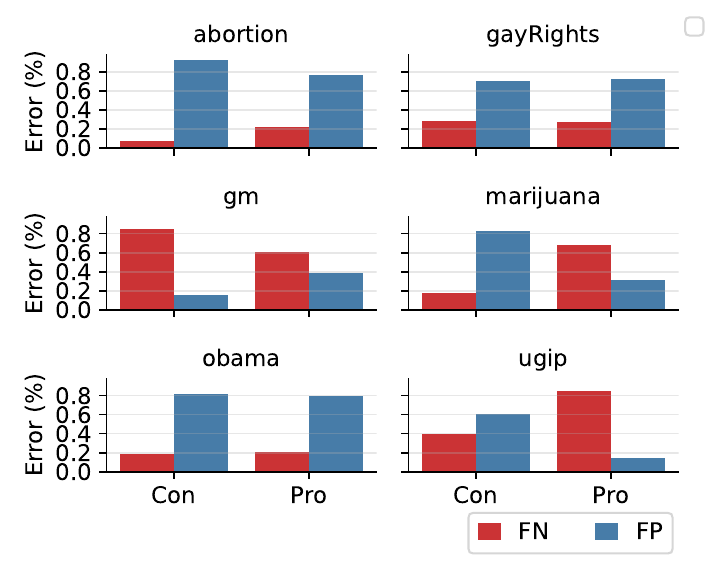}
    \caption{Proportion of false positive and false negative errors for Pro and Con arguments in each dataset.}
    \label{fig:fps}
\end{figure}

\subsection{Error Analysis} 
\label{explanalysis}
%\lea{Structure each of these paragraphs as 1) what you did 2) what you found and 3) why that's an important finding considering downstream applications / bias}

Where exactly did LLMs fail on fine-grained argument detection, extraction and relation classification? To better understand this, we quantitatively and qualitatively inspected the predictions of the overall best k-shot model (Gemini, 5-shot). We systematically compared model predictions against gold labels, analyzing false positives (incorrectly identifying arguments) or false negatives (missing actual arguments) in Task 1, inspecting golden spans and predicted spans in the extraction task (Task 2), and the misclassification patterns in the relationship classification in Task 3.

\paragraph{False positives dominate in argument detection.} As detailed in Figure~\ref{fig:fps}, across the full dataset, false positive predictions (FP) of argument presence significantly outnumber false negatives (FN), accounting for approximately 66\% of all errors. This pattern is particularly strong for Con arguments (which are against a topic), where 76\% of all errors are FPs (62\% for Pro arguments, in support of a topic). In other words: argumentative content is systematically over-predicted in comments that critique a given topic.
% In other words: counter arguments against a topic are systematically over-predicted.
% When broken down by stance, this pattern is even more pronounced for Con arguments, where FPs constitute 76\% of the errors, compared to 62\% for Pro arguments. 

This tendency is particularly strong for the topics \textit{Abortion Rights} and \textit{Obama Presidency}, where FPs for Con arguments account for 92\% and 81\% of errors. The only exception is the topic \textit{Gay Marriage}, where FNs heavily dominate Con arguments.
% (, for topics such as \textit{GM} and \textit{UGIP}, the model often fails to detect Pro arguments, with FNs comprising over 85\% of Pro errors. 
These findings raise concerns for applications like content moderation and debate analysis systems, where stance-specific systematic misclassifications can lead to a skewed picture of opinions as well as erroneous classification of non-argumentative text as supporting particular positions.

\begin{figure}[t]
    \centering
    \includegraphics[width=\linewidth]{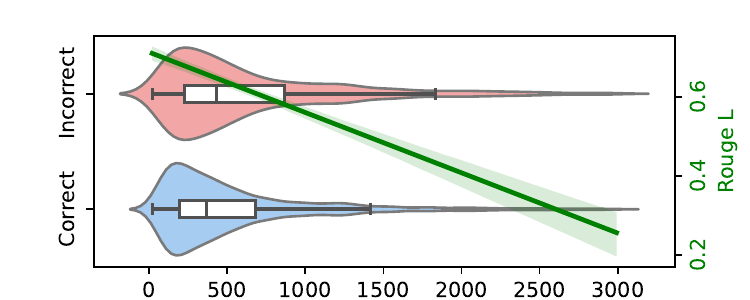}
     \caption{The effect of comment length on comment identification accuracy (Task 1; Violin/box plots) and argument extraction (Task 2; Rouge-L).}
    \label{fig:violin-t1-t2}
\end{figure}

\paragraph{Arguments are harder to identify in long comments.} We observed a significant negative relationship between comment length (up to 3,000 characters to reduce the impact of outliers\footnote{All significance results hold for stricter length thresholds (i.e., even fewer outliers), too, e.g., considering only comments of up to 750 characters.}) and model accuracy across both Task 1 (binary argument detection) and Task 2 (argument extraction). This is illustrated in Figure~\ref{fig:violin-t1-t2}. For Task~1 we find a significant difference in mean length for correctly vs incorrectly classified comments ($t {=} {-}12.103, p {<<} 0.001$). For extraction (Task 2) performance we find a significant negative correlation between comment length and Rouge-L (Pearson’s r ${=} {-}0.27, p {<<} 0.001$). For downstream applications, this length effect could systematically bias system performance against more elaborate reasoning in comments, therefore potentially distorting the representation viewpoints expressed in texts. It also points to an opportunity for future work to address this gap.

\paragraph{Strong and emotional language.} Manual inspection of 50 random mis-classified data points for each task, stratified across topics, revealed systematic language-related patterns in model failures. For Task 1, we observed frequent false positive predictions of arguments in emotionally charged or sarcastic comments (see examples in Table~\ref{tab:error_analysis_task1}). Similar effects were observed for Task 3, where the model most often confused implicit attacks of Pro arguments with explicit attacks in cases where aggressive and offensive rhetoric overshadowed the actual argumentative content (see Table~\ref{tab:fn_errors} for examples). Our findings suggest that strong and emotional language -- which is common in online discussion -- compromises model performance on the identification of argumentative content. Inappropriate reliance on surface-level cues can result in systematic bias in downstream applications.

\begin{table*}
\centering
\begin{small}
\setlength{\tabcolsep}{0.3em}
\begin{tabular}{p{1.1\columnwidth}|p{0.55\columnwidth}|p{0.15\columnwidth}|p{0.15\columnwidth}}
\toprule
\textbf{Comment} & \textbf{Argument} & \textbf{Gold} & \textbf{Pred} \\
\midrule
Immorality should {never has A SAY, should never be accepted} as something normal. Marriage is between a man and a woman not between 2 men or 2 women. It is against our nature, against our God & It is discriminatory to refuse gay couples the right to marry & Implicit Attack & Explicit Attack\\
\midrule
Homosexuality is considered risky behavior and cannot produce offspring and should not be considered with the same respect & Gay couples should be able to take advantage of the fiscal and legal benefits of marriage & Implicit Attack & Explicit Attack\\
\bottomrule
\end{tabular}
\end{small}
\caption{Task 3: Extracts of comments where Gemini incorrectly classified implicit attacks as explicit with strong/emotional language present in the comments.}
\label{tab:fn_errors}
\end{table*}

\section{Conclusion}
We investigated how well LLMs can detect and understand the use of pre-defined arguments in online comments on contested topics. To do so, we separated the objective into three tasks: 1) assessing whether an argument is used in a comment, 2) extracting the exact span in which is it present, 3) and assessing whether the comment supports or attacks the argument. 

We found that overall LLMs perform well on classification tasks (1, 3). While argument span extraction results in terms of Rouge-L appeared weak, manual analysis and additional validation through BERTScore indicates that models often extract argument-relevant spans which, however, may differ from the gold annotations. Task-specific fine-tuning yielded the best results, albeit with considerable computational and environmental costs. Interestingly, increased model size or examples did not consistently boost performance, though LLMs remained robust to example selection.

Our error analysis of one of the strongest LLMs revealed systematic limitations: Gemini systematically over-predicted arguments in emotional content, and performance degrades significantly with comment length. Both calls for follow-up work and raises concerns about reliability for a variety of downstream applications, such as content moderation tools or public opinion analysis where current models could systematically miss long or more nuanced arguments that require extended reasoning. Conversely, Gemini tended to overpredict argumentative content in strongly worded text, indicating overreliance on superficial linguistic cues. Such amplification strongly worded claims by LLMs may pose challenges for balanced, large-scale opinion analysis.

While we split argument analysis into atomic tasks to uncover specific weaknesses, end-to-end models remain appealing. Our results can guide their evaluation by identifying challenge cases for benchmarks and inform design decisions, such as prompt tuning or few-shot selection to address underrepresented arguments

In conclusion, our systematic evaluation provides a thorough overview of current performance, and systematic error analysis. It constitutes a basis for future work to explore how the identified shortcomings can be addressed for instance through improved prompting and fine-tuning, and to broaden our analysis to further topics and genres. 

\section{Limitations}
The data used in this study is limited in scope, both in terms of size and the range of topics and arguments it covers. While this controlled data set enabled a detailed analysis of Large Language Models (LLMs) in argumentation tasks, it may not fully represent the complexity and diversity of real-world argumentation. Notably, the datasets employed were released in 2014, and may not capture more recent arguments or shifts in public opinion. For instance, the arguments related to the subtopic of gay marriage might no longer be relevant, especially given the legalization of gay marriage in the US in 2015, shortly after the data was released. On account of the limited data set size, we needed to conflate all datapoints for Task 1 to fine-tune our RoBERTa baseline. Due to time and cost constraints, as well as environmental considerations, we were only able to fine-tune one LLM (Llama3) on two of the proposed tasks.

\section{Ethical Considerations}
This study investigates the performance of LLMs in AM-related tasks on polarizing topics, which may involve sensitive or controversial discussions. We emphasize that the views in the data do not represent our own views, and that the findings and conclusions of this research are not intended to amplify or legitimize harmful, discriminatory, or unethical viewpoints. Instead, the goal is to evaluate and enhance the understanding of LLMs' capabilities in argument detection, classification and extraction, also analyzing their shortcomings and implications. Our research does not seek to endorse divisive or harmful opinions.

\label{sec:bibtex}
\bibliography{august2024}

\appendix
\section{Lists of Arguments}
\label{sec:appendix}
Here, we present the complete list of pro and con arguments from the original datasets in Table \ref{proconargs}. 

\section{Text Length and Examples}
\label{sec:examples}
This section includes extensive length statistics of the argumentative texts (comments from online discussions) in our data (Table~\ref{tab:text-length-stats}), as well as two examples of such comments (1 for the abortion topic, 1 for the marijuana topic -- Table~\ref{tab:comments-examples}).

\section{Prompts}
\label{app:prompts}

We display the prompts used for our three tasks in Table \ref{app:task1} to Table \ref{app:task3}.

\begin{table*}[ht]
  \begin{small}
  \setlength{\tabcolsep}{0.4em}
    \begin{tabular}{l|p{7cm}|p{7cm}}
    \toprule
    \textbf{Data set} & \textbf{Pro Arguments} & \textbf{Con Arguments} \\
    \midrule
    GM & 
    It is discriminatory to refuse gay couples the right to marry. \newline
    Gay couples should be able to take advantage of the fiscal and legal benefits of marriage. \newline
    Marriage is about more than procreation, therefore gay couples should not be denied the right to marry due to their biology.\newline
    Others &
    Gay couples can declare their union without resort to marriage. \newline
    Gay marriage undermines the institution of marriage, leading to an increase in out-of-wedlock births and divorce rates. \newline
    Major world religions are against gay marriages. \newline
    Marriage should be between a man and a woman.\newline
    Others \\
    \midrule
    UG & 
    Likely to be seen as a state-sanctioned condemnation of religion. \newline
    The principles of democracy regulate that the wishes of American Christians, who are a majority, are honored. \newline
    "Under God" is part of the American tradition and history.\newline
    America is based on democracy and the pledge should reflect the belief of the American majority\newline
    Others &
    Implies ultimate power on the part of the state. \newline
    Removing "under God" would promote religious tolerance. \newline
    Separation of state and religion. \newline
    Others\\
    \midrule
    AB  & 
    Abortion is a woman's right. \newline
    Rape victims need it to be legal. \newline
    A fetus is not a human yet, so it's okay to abort. \newline
    Abortion should be allowed when a mother's life is in danger. \newline
    Unwanted babies are ill-treated by parents and/or not always adopted. \newline
    Birth control fails at times, and abortion is one way to deal with it. \newline
    Abortion is not murder. \newline
    Mother is not healthy/financially solvent. \newline
    Others &
    Put the baby up for adoption. \newline
    Abortion kills a life. \newline
    An unborn baby is a human and has the right to live. \newline
    Be willing to have the baby if you have sex. \newline
    Abortion is harmful to women. \newline
    Others \\
    \midrule
    GR  & 
    Gay marriage is like any other marriage. \newline
    Gay people should have the same rights as straight people. \newline
    Gay parents can adopt and ensure a happy life for a baby. \newline
    People are born gay. \newline
    Religion should not be used against gay rights. \newline
    Others &
    Religion does not permit gay marriages. \newline
    Gay marriages are not normal/against nature. \newline
    Gay parents cannot raise kids properly. \newline
    Gay people have problems and create social issues. \newline
    Others\\
    \midrule
    MA  & 
    Not addictive. \newline
    Used as a medicine for its positive effects. \newline
    Legalized marijuana can be controlled and regulated by the government. \newline
    Prohibition violates human rights. \newline
    Does not cause any damage to our bodies.\newline
    Others &
    Damages our bodies. \newline
    Responsible for brain damage. \newline
    If legalized, people will use marijuana and other drugs more. \newline
    Causes crime. \newline
    Highly addictive. \newline
    Others\\ \\
    \midrule
    OB  & 
    Fixed the economy. \newline
    Ending the wars. \newline
    Better than the Republican candidates. \newline
    Makes good decisions/policies. \newline
    Has qualities of a good leader. \newline
    Ensured better healthcare. \newline
    Executed effective foreign policies. \newline
    Created more jobs. \newline
    Others &
    Destroyed our economy. \newline
    Wars are still ongoing. \newline
    Unemployment rate is high. \newline
    Healthcare bill is a failure. \newline
    Poor decision-maker. \newline
    We have better Republicans than Obama. \newline
    Not eligible as a leader. \newline
    Others\\
    \bottomrule
    \end{tabular}
  \end{small}
  \centering\caption{Pro and Con Arguments for All Subtopics and Datasets}
  \label{proconargs}
\end{table*}

\section{RoBERTa Fine-Tuning}
\label{robfinetun}
We fine-tuned RoBERTa-base using the following configurations for each task:

\begin{itemize}
    \item \textbf{Task 1: Argument Detection}
    \begin{itemize}
        \item Training batch size: 16
        \item Evaluation batch size: 64
        \item Number of epochs: 3
        \item Warmup steps: 500
        \item Weight decay: 0.01
        \item Evaluation strategy: per epoch
        \item Save strategy: per epoch
        \item Load best model at end: Yes
    \end{itemize}

    \item \textbf{Task 2: Argument Extraction}
    \begin{itemize}
        \item Training batch size: 16
        \item Evaluation batch size: 16
        \item Number of epochs: 10
        \item Maximum sequence length: 512
        \item N-best size: 16
        \item Evaluate during training: No
        \item Save checkpoints: No
        \item Overwrite output directory: Yes
        \item Save model every epoch: No
    \end{itemize}

    \item \textbf{Task 3: Relationship Classification}
    \begin{itemize}
        \item Training batch size: 16
        \item Evaluation batch size: 64
        \item Number of epochs: 3
        \item Warmup steps: 500
        \item Weight decay: 0.01
        \item Evaluation strategy: per epoch
        \item Save strategy: per epoch
        \item Load best model at end: Yes
        \item Optimization metric: F1
        \item Optimization goal: maximize
    \end{itemize}
\end{itemize}

All models were trained on a single NVIDIA V100 GPU using the RoBERTa-base checkpoint as the initial model.

\section{Parameter-efficient finetuning (PEFT) of LlaMA}
\label{app:lora}

For PEFT, we used an implementation of low-rank adaptation (LoRA) from Unsloth AI\footnote{\url{https://github.com/unslothai/unsloth}} with the following hyperparameters:
\begin{itemize}
\item load in 4 bit = False
\item r = 16
\item target modules = q\_proj, k\_proj, v\_proj, o\_proj, gate\_proj, up\_proj, down\_proj
\item lora alpha = 16
\item lora dropout = 0
\item bias = none
\item use gradient checkpointing = unsloth
\item use rslora (rank stabilized LoRA) = False
\end{itemize}

The finetuning was performed with 5-fold cross-validation (data split of 60-20-20 for train-dev-test sets, with test splits covering the whole dataset). For the classification task, the splits were stratified. The training used 8-bit Adam as optimizer and the standard learning rate of 2e-4. The number of training steps was proportional to the data size, with loss falling to near-zero values as a stop signal, and roughly amounted to 3 full epochs for the classification task and 5 full epochs for the span extraction task.

\section{Text Length and Examples}
\label{sec:examples}
This section includes extensive length statistics of the argumentative texts (comments from online discussions) in our data (Table~\ref{tab:text-length-stats}), as well as two examples of such comments (1 for the abortion topic, 1 for the marijuana topic -- Table~\ref{tab:comments-examples}).

\begin{table*}[ht]
\setlength{\tabcolsep}{0.4em}
\centering
\begin{tabular}{l c c c c}
\hline
Topic & Min Characters & Max Characters & Mean Characters & Median Characters \\
\hline
Gay Marriage & 33 & 2,454 & 683.06 & 672.0 \\
UGIP & 31 & 1,317 & 486.21 & 405.0 \\
Gay Rights & 44 & 6,441 & 772.25 & 473.0 \\
Abortion & 33 & 23,055 & 981.52 & 536.0 \\
Marijuana & 21 & 3,658 & 731.44 & 495.0 \\
Obama & 53 & 14,904 & 846.31 & 434.0 \\
\hline
\end{tabular}
\caption{Text Length Statistics of comments across topics}
\label{tab:text-length-stats}
\end{table*}

\begin{table*}[ht]
\setlength{\tabcolsep}{0.4em}
\centering
\begin{tabular}{l p{0.7\textwidth}}
\hline
Topic & Comment \\
\hline
Abortion & Why should you kill a innocent baby? That is exactly what abortion is. Even though the mother does not want the baby, she should still have it. Most of the people who want an abortion and never go through with it, actually say they would regret killing the baby. Should America become `"I get to do whatever I want to just because I can"? \\
\hline
Marijuana & I believe marijuana should be legal for many reasons. First of all it is proven that it helps with different things medically such as when going through chemo it gives you appetite, it helps with pain control etc. Also i feel personally that alcohol is more dangerous then marijuana. I have seen many people killed from drunk drivers and it is a shame that so many people drive drunk. But, i have never heard of anyone dying from smoking too much weed, killing someone from an accident because they smoked weed, or anything like that.. Marijuana is a natural herb and it is legal in many other places and could possible make some money for the country if legalized! \\
\hline
\end{tabular}
\caption{Example Comments for Abortion and Marijuana Topics}
\label{tab:comments-examples}
\end{table*}

%We report the prompts used for each task here. Additional to asking for JSON output in the prompts themselves, we use the BaseModel class from the library Pydantic\footnote{\url{https://docs.pydantic.dev/latest/api/base_model/}} to ensure JSON outputs. When running these prompts in a few-shot setting, we instantiate the model with the same examples in each task. \\

\section{Prompts}
\label{app:prompts}

We display the prompts used for our three tasks in Table \ref{app:task1} to Table \ref{app:task3}.

\begin{table*}[h]
\begin{tabular}{p{\textwidth}}
\toprule
Analyze whether the following comment about \{topic\} contains a specific argument.\\

Argument to check for: \{argument\}\\

Instructions:\\
1. Determine if the comment explicitly or implicitly uses the given argument\\
2. Assign a binary label:\\
- 1 if the argument is present\\
- 0 if the argument is not present\\

Requirements:\\
- Only use 1 or 0 as labels\\
- Provide output in valid JSON format\\
- Do not repeat or include the input text in the response\\
- Focus solely on the presence/absence of the specific argument\\

Return your analysis in this exact JSON format:\\
\texttt{{"id": "{id}", "label": label\_value}} \\

Analyze the following comment in relation to the given argument:\\
\bottomrule
\end{tabular}
\caption{Prompt for Task 1}
\label{app:task1}
\end{table*}

\begin{table*}[h]
\begin{tabular}{p{\textwidth}}
\toprule
            Task: Text Span Identification for Arguments about \{topic\}\\
            Target Argument: \{argument\_text\}\\
            Role: You are an expert in argument analysis and logical reasoning,\\ specializing in identifying rhetorical patterns in social discourse.\\

            Step-by-Step Instructions:\\
            1. Read the input text carefully\\
            2. Locate exact text spans that:\\
            - Directly reference the target argument\\
            - Express the same idea as the argument\\
            3. Extract the precise text span\\
            4. Format the output according to specifications\\

            Critical Requirements:\\
            - Extract EXACT text only (no paraphrasing)\\
            - Include COMPLETE relevant phrases\\
            - Use MINIMUM necessary context\\
            - Maintain ORIGINAL formatting\\
            - Return VALID JSON only\\

            Output Schema:\\
            \{
                "id": "\{id\}",\\
                "span": "exact\_text\_from\_comment"  \# must be verbatim quote\\
            \}
            
            Input Text: \\
\bottomrule
\end{tabular}
\caption{Prompt for Task 2}
\label{app:task3} 
\end{table*}

\begin{table*}[h]
\begin{tabular}{p{\textwidth}}
\toprule
            Task: Binary Classification of Arguments about \{topic\}\\
            Input Text: \{comment\_text\}\\
            Target Argument: \{argument\_text\}\\
            Role: You are an expert in argument analysis and logical reasoning,\\ specializing in identifying rhetorical patterns in social discourse.\\

            Step-by-Step Instructions:\\
            1. Read the input text thoroughly\\
            2. Evaluate the text's relationship to the target argument, examining:\\
            - Direct support or opposition\\
            - Implicit agreement or disagreement\\
            3. Make a binary classification decision\\
            4. Format the output according to specifications\\

            Classification Rules:\\
            - Label = 5: Comment supports/agrees with argument\\
            - Label = 1: Comment attacks/disagrees with argument\\

            Critical Requirements:\\
            - Use ONLY specified labels (1 or 5)\\
            - Do NOT quote or repeat input texts\\
            - Return VALID JSON only\\

            Output Schema:
            \{
                "id": "\{id\}",
                "label": label\_value  \# must be 1 or 5 without quotes
            \}\\
            
            Input Text: \\
\bottomrule
\end{tabular}
\caption{Prompt for Task 3 - Binary}
\label{app:task1}
\end{table*}

\begin{table*}[h]
\begin{tabular}{p{\textwidth}}
\toprule
            Task: Classification of Arguments about \{topic\}\\
            Input Text: \{comment\_text\}\\
            Target Argument: \{argument\_text\}\\
            Role: You are an expert in argument analysis and logical reasoning,\\ specializing in identifying rhetorical patterns in social discourse.\\

            Step-by-Step Instructions:\\
            1. Read the input text thoroughly\\
            2. Evaluate the text's relationship to the target argument, examining:\\
            - Direct support or opposition\\
            - Implicit agreement or disagreement\\
            3. Make a binary classification decision\\
            4. Format the output according to specifications\\

            Classification Rules:\\
            - Label = 5: Comment supports/agrees with argument\\
            - Label = 4: Comment supports/agrees with argument implicitly/indirectly\\
            - Label = 2: Comment attacks/disagrees with argument implicitly/indirectly\\
            - Label = 1: Comment attacks/disagrees with argument\\

            Critical Requirements:\\
            - Use ONLY specified labels (1 or 5)\\
            - Do NOT quote or repeat input texts\\
            - Return VALID JSON only\\

            Output Schema:
            \{
                "id": "\{id\}",
                "label": label\_value  \# must be 1, 2, 4 or 5 without quotes
            \}\\
            
            Input Text: \\
\bottomrule
\end{tabular}
\caption{Prompt for Task 3 - Full Scale}
\label{app:task1}
\end{table*}

\section{RoBERTa Fine-Tuning}
\label{robfinetun}
We fine-tuned RoBERTa-base using the following configurations for each task:

\begin{itemize}
    \item \textbf{Task 1: Argument Detection}
    \begin{itemize}
        \item Training batch size: 16
        \item Evaluation batch size: 64
        \item Number of epochs: 3
        \item Warmup steps: 500
        \item Weight decay: 0.01
        \item Evaluation strategy: per epoch
        \item Save strategy: per epoch
        \item Load best model at end: Yes
    \end{itemize}

    \item \textbf{Task 2: Argument Extraction}
    \begin{itemize}
        \item Training batch size: 16
        \item Evaluation batch size: 16
        \item Number of epochs: 10
        \item Maximum sequence length: 512
        \item N-best size: 16
        \item Evaluate during training: No
        \item Save checkpoints: No
        \item Overwrite output directory: Yes
        \item Save model every epoch: No
    \end{itemize}

    \item \textbf{Task 3: Relationship Classification}
    \begin{itemize}
        \item Training batch size: 16
        \item Evaluation batch size: 64
        \item Number of epochs: 3
        \item Warmup steps: 500
        \item Weight decay: 0.01
        \item Evaluation strategy: per epoch
        \item Save strategy: per epoch
        \item Load best model at end: Yes
        \item Optimization metric: F1
        \item Optimization goal: maximize
    \end{itemize}
\end{itemize}

All models were trained on a single NVIDIA V100 GPU using the RoBERTa-base checkpoint as the initial model.

\section{Parameter-efficient finetuning (PEFT) of LlaMA}
\label{app:lora}

For PEFT, we used an implementation of low-rank adaptation (LoRA) from Unsloth AI\footnote{\url{https://github.com/unslothai/unsloth}} with the following hyperparameters: 

\begin{itemize}
\item load in 4 bit = False
\item r = 16
\item target modules = q\_proj, k\_proj, v\_proj, o\_proj, gate\_proj, up\_proj, down\_proj
\item lora alpha = 16
\item lora dropout = 0
\item bias = none
\item use gradient checkpointing = unsloth
\item use rslora (rank stabilized LoRA) = False
\end{itemize}

The finetuning was performed with 5-fold cross-validation (data split of 60-20-20 for train-dev-test sets, with test splits covering the whole dataset). For the classification task, the splits were stratified. The training used 8-bit Adam as optimizer and the standard learning rate of 2e-4. The number of training steps was proportional to the data size, with loss falling to near-zero values as a stop signal, and roughly amounted to 3 full epochs for the classification task and 5 full epochs for the span extraction task. 

The same prompts and example/label formats were used for finetuning as for the zero-shot and few-shot experiments (see Appendix~\ref{app:prompts}).

\end{document}